\documentclass[manuscript,nonacm]{acmart}

\usepackage{booktabs}
\usepackage{multirow}
\usepackage{tikz}
\usetikzlibrary{arrows.meta,positioning}

\title{Textual User Taste: Natural-Language User Context for Foundation-Model
Recommender System at Scale}

\author{Ghazal Fazelnia}
\author{Paul Gigioli}
\author{Eliza Klyce}
\author{Sharon Zheng}
\author{Katie Zelvin}
\author{Ye Myat Thein}
\author{Anurag Deshpande}
\author{Seda Davtyan}
\author{Kate Remeika}
\author{Maya Hristakeva}
\author{Erik Franco}
\author{Karen Banzon}
\author{Peng Ge}
\author{Jacqueline Wood}
\author{Nandini Singh}
\author{David Murgatroyd}
\author{Mounia Lalmas}
\author{Yves Raimond}
\author{Andreas Damianou}

\affiliation{%
  \institution{Spotify}
  \country{USA, UK, Sweden}
}

\begin{document}

\begin{abstract}
Foundation model recommender systems require user context that can be
consumed by large language models, reasoned over, and refined through
natural-language interaction. Traditional behavioral embedding vectors remain highly effective for retrieval and ranking, but they are opaque to users and not natively expressed for language model workflows. We present \emph{Textual User Taste}, a system that generates structured natural-language taste profiles from listening behavior, interaction signals, content metadata, and optional user feedback, and deploys them to millions of Spotify users. We describe the end-to-end production lifecycle required to generate, evaluate, optimize, and maintain these representations at industrial scale, including prompt development and compression, user steering, and integration with downstream personalization systems.

Because no unique ground-truth taste profile exists, we introduce a multi-faceted evaluation framework to evaluate taste profiles as a production representation: they carry user-specific predictive signal independently, and when integrated with behavioral embeddings, improve MRR by 0.6\% for future-track prediction and NDCG@7 by 2.2\% for search ranking. Our evaluation also reveals that taste profiles support positive natural-language steering, while exposing important limitations, including challenges with negation and short-term temporal adaptation. These findings position taste profiles not as replacements for behavioral embeddings, but as an interpretable and steerable interface between evolving user context and foundation-model recommender systems.
\end{abstract}

\begin{CCSXML}
<ccs2012>
 <concept>
  <concept_id>10002951.10003260.10003277</concept_id>
  <concept_desc>Information systems~Recommender systems</concept_desc>
  <concept_significance>500</concept_significance>
 </concept>
 <concept>
  <concept_id>10010147.10010178.10010224</concept_id>
  <concept_desc>Computing methodologies~Natural language processing</concept_desc>
  <concept_significance>300</concept_significance>
 </concept>
</ccs2012>
\end{CCSXML}
\ccsdesc[500]{Information systems~Recommender systems}
\ccsdesc[300]{Computing methodologies~Natural language processing}

% \keywords{user modeling, textual user profiles, context-aware recommendation,
% foundation models, personalization}
% \authorsaddresses{}
\authorsaddresses{Corresponding author: Ghazal Fazelnia, ghazalf@spotify.com}
\maketitle
  \fancypagestyle{copyrightpage}{%
    \fancyhf{}
    \fancyfoot[L]{\footnotesize Copyright held by the author(s)}
    \fancyfoot[R]{\thepage}
  }
  \thispagestyle{copyrightpage}

\section{Introduction}

User modeling in recommender systems has traditionally relied on latent
representations learned from interaction data, from matrix factorization~\citep{koren2009matrix} to neural collaborative filtering~\citep{he2017neural} and large-scale sequential models~\citep{pancha2022pinnerformer,zhai2024hstu}. These representations are compact, efficient, and highly predictive. However, they are also opaque: users cannot inspect a behavioral embedding, correct an incorrect assumption about their preferences, or express a new contextual preference.

Foundation models change the role of user representations in recommender
systems. Conversational recommenders, natural-language search, and agentic
content discovery consume \emph{textual} context, reason about intent, and generate
language as part of the recommendation process. Passing raw interaction
history to every downstream model is expensive and often exceeds practical
context budgets. A textual taste profile provides a different abstraction: a
cached, semantically organized representation of a user's longer-term preferences and
consumption patterns that language models can readily consume and that users can inspect
and refine. More broadly, it extends context-aware recommendation~\citep{adomavicius2011cars} by expressing behavioral, temporal, and explicit user context as reusable natural-language representations.

Recent research has demonstrated the value of language-based user profiles for
recommendation, transparency, and control~\citep{zhou2024lfm,ramos2024transparent,sguerra2025biases}. More recent work has optimized profile generation using downstream or self-supervised behavioral
objectives~\citep{arora2026highfidelity,ju2026bump,chen2026duet}, while
industrial systems have demonstrated real-time persona generation at very
large scale~\citep{wang2026personas}. These advances establish textual user profiles as a viable user representation for recommender systems. However, they also raise a broader lifecycle question: how should such representations be generated, evaluated, refreshed, and governed once they become shared infrastructure across multiple products?

Our central argument is that taste profiles should be evaluated as operational user representations rather than simply as pieces of text. High-quality language is necessary but not sufficient: a useful profile must remain grounded in observed behavior, distinguish one user from another, influence downstream systems in intended ways, and respond predictably to explicit user feedback.
We therefore treat generation and evaluation as a closed operational loop in which evaluation continuously informs how taste profiles are generated, optimized, deployed, and monitored.

This paper studies that lifecycle through \emph{Textual User Taste}, Spotify's production system for generating and serving textual taste profiles, surfaced in a user-facing experience and consumed by both foundation-model and conventional personalization systems. The taste profiles summarize listening patterns across content types from behavioral signals and optional user-authored feedback. Operating this representation at scale exposes challenges that extend beyond taste profile generation: there is no single reference taste profile; fluent text may be insufficiently grounded in observed behavior; prompt changes propagate across multiple downstream consumers; the same inputs may produce different profiles across regenerations; and taste profiles may expose unsupported or sensitive inferences about users.

Our contributions are:
\begin{itemize}
  \item We present an end-to-end production framework for generating, evaluating, maintaining, and serving structured textual user representations at scale.
  \item We introduce a comprehensive evaluation framework for taste profiles that combines intrinsic and functional evaluation to address the absence of a unique ground-truth profile.
  \item We provide an empirical study of the strengths and limitations of taste profiles in production recommender systems, identifying where they complement behavioral embeddings and where important challenges remain.
\end{itemize}

\section{Related Work}

\paragraph{Textual user profiles.}
The Language-based Factorization Model (LFM) generates compact profiles from rating histories for downstream prediction, showing benefits in cold-start settings \citep{zhou2024lfm}. Natural-language profiles have also been shown to support scrutable recommendation, allowing users to inspect and edit their profiles to influence recommendations \citep{ramos2024transparent}. TEARS aligns textual and collaborative representations while exposing an editable profile \citep{penaloza2025tears}. PURE incrementally updates profiles from reviews, while PersonaX maintains multi-interest profiles for recommendation agents \citep{bang2025pure,shi2025personax}. In the music domain, recent work has examined whether listeners identify with LLM-generated taste profiles and how profile quality varies across users and items \citep{sguerra2025biases}. Collectively, these studies establish the value of textual user profiles, but primarily study profile generation or use within a single domain, task, or experimental setting rather than persistent context shared across production systems. Our work addresses this gap by studying taste profiles as production infrastructure, where the absence of a unique reference profile makes evaluation, maintenance, and deployment central challenges.

\paragraph{Industrial scale and complementary representations.}
Recent work has demonstrated scalable generation of real-time user personas for industrial recommendation systems using distillation, asynchronous inference, and clustered input representations \citep{wang2026personas}. While this establishes the feasibility of large-scale persona generation, it does not consider a persistent, user-visible, multi-content representation that can be explicitly refined through natural-language feedback and reused across multiple downstream systems. A related line of work injects collaborative embeddings into language models or aligns semantic and collaborative representations \citep{ren2024rlmrec,zhang2025collm,liu2024llmesr}. These approaches propose new mechanisms for combining textual and behavioral representations, but largely leave open how such representations should be evaluated, maintained, and operated as production context. Rather than proposing a new fusion architecture, we treat taste profiles and behavioral embeddings as complementary production representations and study where taste profiles provide value beyond behavioral embeddings alone.

\section{Textual User Taste System}

This section describes the Textual User Taste system and the production lifecycle of its taste profiles. We present the key components of the system, from aggregating behavioral signals and generating structured profiles to serving them across downstream consumers and continuously refining them through evaluation and user feedback.

\subsection{Inputs and Profile Structure}

For each user, the system generates a structured taste profile from aggregated behavioral signals rather than raw interaction histories. Inputs span multiple temporal horizons and include artists, tracks, genres, podcasts, audiobooks, topical interests, follows, negative feedback (e.g., hides), discovery patterns (e.g., exploration versus repeat listening), and time-of-week consumption. Content metadata, including titles, descriptions, categories, and creator names, provide additional semantic grounding. Using aggregated signals reduces prompt size, lowers generation cost, and eliminates the need to pass complete interaction histories to the generator, while also capturing rich semantic information across multiple levels of granularity. An optional user-authored note (e.g., ``more jazz'') provides explicit feedback; notes are filtered for relevance and safety and kept separate from observed behavior so the generator can distinguish stated from inferred preferences.

The generator returns a structured taste profile with four fields:
\begin{itemize}
  \item \textit{Music Taste}: enduring musical preferences, affinities,
  and representative content;
  \item \textit{Podcast Taste}: preferred topics, formats, and engaged shows;
  \item \textit{Audiobook Taste}: preferred genres, authors, and engagement patterns;
  \item \textit{Listening Rhythms}: recurring temporal and cross-content consumption patterns.
\end{itemize}
This schema separates enduring content preferences from consumption context, allowing downstream systems to select only the fields relevant to a given task while avoiding unsupported inferences when behavioral evidence is sparse.

\subsection{Generation, Guardrails, and Serving}

The structured input is supplied to a cost-efficient instruction-tuned LLM that generates the taste profile.
The prompt explains the meaning of each input feature, specifies the output schema and style,
prioritizes recent evidence while retaining durable preferences, and instructs the model to avoid
unsupported entities and sensitive inferences. Guardrails are applied both before and after generation.
Input safeguards filter irrelevant, unsafe, or adversarial user notes, while output safeguards monitor baseline safety together with profile-specific risks, including protected-attribute inference, personally identifiable information, emotional overreach, and unsupported claims about creators.

The model and prompt are versioned together, allowing candidate versions to be evaluated before deployment and monitored after rollout. Taste profiles are stored and served to both user-facing and backend consumers rather than regenerated for every request, reducing inference cost while providing downstream systems with a stable representation. When a user submits an accepted note, it is incorporated during a subsequent regeneration, making profile updates part of a controlled feedback-and-refresh loop rather than unrestricted editing of stored text.

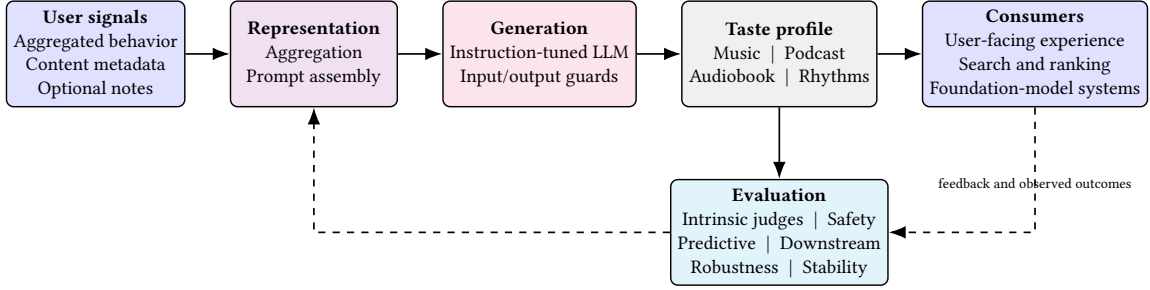
\begin{figure}[t]
  \centering
  \resizebox{\linewidth}{!}{%
  \begin{tikzpicture}[
    node distance=0.65cm,
    box/.style={draw, rounded corners=3pt, align=center, minimum height=1.55cm,
      minimum width=2.45cm, font=\small, line width=0.6pt},
    flow/.style={-{Latex[length=2.5mm]}, line width=0.7pt},
    feedback/.style={-{Latex[length=2.5mm]}, line width=0.7pt, dashed}
  ]
    \node[box, fill=blue!12] (input) {\textbf{User signals}\\Aggregated behavior\\Content metadata\\Optional notes};
    \node[box, fill=violet!12, right=of input] (prep) {\textbf{Representation}\\Aggregation\\Prompt assembly};
    \node[box, fill=magenta!12, right=of prep] (gen) {\textbf{Generation}\\Instruction-tuned LLM\\Input/output guards};
    \node[box, fill=gray!12, right=of gen] (profile) {\textbf{Taste profile}\\Music \;\textbar\; Podcast\\Audiobook \;\textbar\; Rhythms};
    \node[box, fill=blue!12, right=of profile] (use) {\textbf{Consumers}\\User-facing experience\\Search and ranking\\Foundation-model systems};
    \node[box, fill=cyan!10, below=1.05cm of profile] (eval) {\textbf{Evaluation}\\Intrinsic judges \;\textbar\; Safety\\Predictive \;\textbar\; Downstream\\Robustness \;\textbar\; Stability};

    \draw[flow] (input) -- (prep);
    \draw[flow] (prep) -- (gen);
    \draw[flow] (gen) -- (profile);
    \draw[flow] (profile) -- (use);
    \draw[flow] (profile) -- (eval);
    \draw[feedback] (eval.west) -| (prep.south);
    \draw[feedback] (use.south) |- node[pos=0.25, below, font=\scriptsize]{feedback and observed outcomes} (eval.east);
  \end{tikzpicture}}
  \caption{Textual User Taste lifecycle. Aggregated behavior, metadata, and
  optional user feedback are converted into a structured prompt. The generated
  fields are evaluated and served to user-facing and downstream personalization
  systems; feedback and evaluation results inform later generations.}
  \label{fig:system-overview}
\end{figure}

Figure~\ref{fig:system-overview} summarizes the complete lifecycle. Evaluation is an integral part of the system rather than a final validation step: signals from LLM judges, predictive tasks, robustness tests, and downstream evaluations continuously inform prompt refinement, model selection, and future profile generations.

\subsection{Prompt Development and Optimization}

Prompt development was treated as an iterative optimization process balancing profile quality and production cost. We iteratively refined seven prompt versions differing in instruction specificity, examples, tone, and safety guidance, assessing each with LLM judges on factuality, coherence, and specificity. Across iterations, factuality improved from 7.70 to 8.57 and coherence from 6.46 to 7.98 on a ten-point scale. The longest prompt achieved the highest factuality and coherence, whereas a shorter variant produced the best specificity, showing that prompt length alone is a poor proxy for quality.

We explored two complementary optimization approaches. First, LLMLingua \citep{jiang2023llmlingua} compressed prompts by removing lower-information instruction tokens while preserving the schema and safety instructions. Table~\ref{tab:compression} shows a clear trade-off: two-fold compression preserved factuality and structured validity, whereas more aggressive compression substantially degraded both. Second, we formulated prompt construction as a program optimization problem using DSPy MIPROv2 and BootstrapFewShot \citep{khattab2024dspy}. The resulting prompt achieved an 80.2\% composite objective (average of the evaluation metrics) using approximately 2,000 tokens (i.e., one-third of the length of the best hand-authored prompt).

These approaches generate candidate prompts. Whether a prompt is deployed depends on the broader evaluation framework described in the next section, ensuring that optimization is guided by downstream utility rather than intrinsic quality alone.

\begin{table}[t]
  \caption{Instruction compression trade-offs. Factuality is scored from 1 to
  10.}
  \label{tab:compression}
  \centering
  \begin{tabular}{lrrr}
    \toprule
    \textbf{Condition} & \textbf{Tokens} & \textbf{Factuality} &
    \textbf{Valid output} \\
    \midrule
    Uncompressed & $\sim$6,800 & 8.66 & 100\% \\
    LLMLingua $2\times$ & 3,428 & 8.61 & 100\% \\
    LLMLingua $4\times$ & 1,803 & 7.80 & 73.1\% \\
    LLMLingua $8\times$ & 1,202 & 3.31 & 16.5\% \\
    \bottomrule
  \end{tabular}
\end{table}

\section{Evaluation Methodology}
\label{sec:evaluation}

Because taste profiles have no unique ground-truth representation, we evaluate them through complementary intrinsic and functional assessments.

\subsection{Intrinsic Quality and Robustness}

Because taste profiles have no unique ground-truth representation, intrinsic evaluation must assess multiple aspects of quality rather than rely on a single score. We therefore use seven LLM judges to evaluate factuality, coherence, coverage, specificity, diversity, relevance, and safety. These dimensions capture whether a profile is grounded, informative, well formed, and appropriate for downstream use. Each judge follows a task-specific rubric and receives only the evidence required for its assessment, allowing unsupported inferences to be distinguished from poor writing.

We use this evaluation suite to compare prompt variants, generator models, and prompt compression strategies, while factuality and safety assessments also support production monitoring. Intrinsic quality alone, however, cannot establish whether a profile benefits downstream systems. We therefore complement these measures with the functional evaluations described in the following subsection. LLM-judge scores are used as diagnostic signals rather than definitive measures of profile quality because they may inherit model biases and exhibit inconsistent calibration across languages.

Because users can explicitly steer their taste profiles, we also evaluate how the system handles challenging feedback scenarios. We therefore constructed approximately 50 stress scenarios spanning nine categories, including contradictory behavior and explicit feedback, prompt injection, personally identifiable information, unsafe content, multilingual inputs, shared-account behavior, empty profiles, and requests that cannot be faithfully represented. These scenarios evaluate both whether generated profiles remain safe and whether conflicts between observed behavior and explicit user feedback are handled transparently.

\subsection{Behavioral and Downstream Tests}

We complement intrinsic evaluation with four functional evaluations, each designed to assess a different aspect of taste profiles: predictive value, user specificity, downstream utility, and steerability.

{\it{Future-track prediction.}} We evaluate whether taste profiles provide complementary predictive signal beyond behavioral embeddings. We compare four user representations in the same downstream ranking task, training the same ranking model for each one: a production behavioral embedding \citep{fazelnia2025generalized}, a taste-profile embedding, their concatenation, and a gated fusion of the two. Taste-profile embeddings are set to 120 dimensions using a Matryoshka text encoder to match the dimensionality of the behavioral embeddings. The evaluation uses 400,000 user--track examples, split 80/20 by user. Results are averaged over five runs and reported using MRR, ROC-AUC, and NDCG@10.

{\it{User specificity.}} We evaluate whether taste profiles capture user-specific information that generalizes beyond observed listening behavior. For each of the users in the sample set, an LLM ranks five candidate tracks consisting of one track the user consumed in a held-out future period and four distractors. We compare three conditions: no profile, the user's own profile, and a randomly assigned profile serving as a control.

{\it{Search ranking.}} We evaluate whether taste profiles improve downstream search personalization. A language model is fine-tuned for search-result ranking under four user-context configurations: no user representation, taste profile only, behavioral user embedding only, and both representations together. The evaluation contains 10,000 query sessions and approximately 5.6 million query--candidate pairs. Performance is measured using SessionNDCG@7 on two temporal train--test partitions.

{\it{Steering.}} We evaluate whether natural-language feedback steers recommendations in the requested direction. For a sample of users and candidate tracks, we augment each taste profile with requests such as ``more jazz'', recompute the profile embedding, and measure whether resulting recommendations shift toward the requested genre or mood. We quantify this shift using semantic similarity and Top-20 precision with respect to the requested genre or mood. We additionally evaluate negative requests (e.g., ``less electronic music''), situational requests (e.g., ``music for deep focus''), and multi-turn refinements in which users iteratively refine their preferences. This is a controlled representation study that evaluates whether textual feedback produces the intended changes in an embedding-based recommender rather than measuring user satisfaction.

\subsection{Representation Complementarity and Temporal Stability}

{\it{Representation complementarity.}} We evaluate the extent to which taste profiles relate to behavioral embeddings. For a sample of users, we compare pairwise similarity rankings, nearest-neighbor relationships, and cluster assignments between the two representation spaces. Structural agreement is quantified using Representational Similarity Analysis (RSA) with Spearman correlation, nearest-neighbor overlap at ($k=5$), and Adjusted Rand Index (ARI). These metrics characterize the structural relationship between the two representation spaces rather than quantifying the amount of information contained in either representation.

{\it{Temporal stability.}}   Temporal stability is measured by regenerating taste profiles after three and seven days under largely unchanged behavioral inputs. For a sample of users, profiles are regenerated using data from a reference day and again after three and seven days. We compare changes in the inputs and seven output measures, including embedding cosine similarity and surface-form similarity. Because the profiles summarize long-term listening behavior, this experiment primarily measures generation stability under nearly unchanged user behavior rather than responsiveness to genuine preference change.

\section{Results}
\label{sec:results}

This section evaluates taste profiles from four complementary perspectives, corresponding to the evaluation framework introduced in Section 4. 
Intrinsic evaluation confirms that profile quality is inherently multidimensional. Principal component analysis of the seven intrinsic quality scores identifies two dominant components explaining 93.7\% of the variance: one associated with semantic richness and another with factual quality. Moreover, factuality and relevance are only weakly negatively correlated ($r=-0.18$), indicating that no single intrinsic metric adequately characterizes profile quality. This finding supports our use of multiple complementary quality dimensions throughout the evaluation.

\subsection{Text Adds Complementary Predictive Signal}

Table~\ref{tab:learned-ranking} shows that behavioral embeddings remain the
strongest standalone representation for future-track prediction. Taste
profiles alone achieve competitive performance, but consistently underperform
behavioral embeddings across all three metrics. Combining the two
representations yields the best overall performance: concatenation improves MRR
from 0.956 to 0.962, ROC-AUC from 0.957 to 0.963, and NDCG@10 from 0.932 to
0.941. The ROC-AUC improvement over the behavioral baseline is statistically
significant, whereas gated fusion performs similarly but offers no
further improvement. These results indicate that taste profiles
provide complementary predictive signal rather than replacing behavioral
embeddings. The benefit is not universal, however: approximately 30\% of users
achieve better predictions than with the behavioral embedding alone.

\begin{table}[t]
  \centering

  \begin{minipage}[t]{0.56\linewidth}
    \vspace{0pt}
    \centering
    \caption{Future-track ranking (five-run mean).}
    \label{tab:learned-ranking}
    \small
    \setlength{\tabcolsep}{3pt}
    \renewcommand{\arraystretch}{0.95}
    \begin{tabular}{@{}lrrr@{}}
      \toprule
      \textbf{Representation} & \textbf{MRR} & \textbf{AUC} &
      \textbf{NDCG@10} \\
      \midrule
      Behavioral & 0.956 & 0.957 & 0.932 \\
      Taste profile & 0.937 & 0.946 & 0.912 \\
      Taste profile + behavioral & \textbf{0.962} & \textbf{0.963} &
        \textbf{0.941} \\
      Gated fusion & 0.961 & 0.962 & 0.940 \\
      \bottomrule
    \end{tabular}
  \end{minipage}
  \hfill
  \begin{minipage}[t]{0.42\linewidth}
    \vspace{0pt}
    \centering
    \caption{Search ranking; uplift is relative to no context.}
    \label{tab:search}
    \small
    \setlength{\tabcolsep}{3pt}
    \renewcommand{\arraystretch}{0.95}
    \begin{tabular}{@{}lrr@{}}
      \toprule
      \textbf{Context} &
      \shortstack{\textbf{NDCG}\\\textbf{@7}}
      \textbf{Uplift} \\
      \midrule
      % None & 0.6242 & --- \\
      % Taste profile & 0.6296 & +0.9\% \\
      % Behavioral & 0.6345 & +1.7\% \\
      % Combined & \textbf{0.6380} & \textbf{+2.2\%} \\
      None  & baseline \\
      Taste profile &  +0.9\% \\
      Behavioral &  +1.7\% \\
      Combined & \textbf{+2.2\%} \\
      \bottomrule
    \end{tabular}
  \end{minipage}
\end{table}

The user-specificity evaluation tests whether the gains observed above arise
from information specific to an individual user rather than simply from
providing additional prompt context. Using the user's own taste profile increases MRR
from 0.656 to 0.776 (+18.3\%; $p=0.0003$, $d=0.26$), whereas replacing it with
another user's taste profile reduces MRR to 0.477. This demonstrates that the LLM
benefits from the user-specific information encoded in the profile rather than
from the presence of additional text alone. While this does not imply that every
statement in the generated profile is factually correct, it does show that the profile captures user-specific information that is predictive of future listening.

\subsection{Downstream Context Works Best in Combination}

Table~\ref{tab:search} evaluates whether taste profiles improve a real
downstream personalization task. A taste profile alone improves SessionNDCG@7 by
0.9\% over the no-user-context baseline, whereas the behavioral embedding improves
performance by 1.7\%. Combining the two representations yields the largest
improvement (2.2\%). Across a second temporal partition, the standalone profile and
behavioral representations exhibit inconsistent gains, whereas their
combination remains consistently positive. These results reinforce the central finding of the previous subsection: taste profiles complement rather than replace behavioral embeddings. Their greatest value lies in providing complementary context for downstream personalization rather than serving as standalone user representations.

% Taste profiles also influence downstream LLM-based recommendation applications. We evaluate an LLM-based home-page generator that assembles personalized recommendation pages conditioned on the user's taste profile. Changing the taste profile produces measurable differences in the generated recommendations across multiple evaluation dimensions, including taste alignment, freshness, coherence, and generation errors. Different profile variants perform best on different downstream evaluation dimensions, indicating that no single profile design is optimal across all criteria. These results should be interpreted as sensitivity analyses rather than evidence of improved user outcomes. They nevertheless show that changes to the taste profile directly influence downstream recommendation generation, reinforcing the importance of evaluating taste profiles through the systems that consume them.

\subsection{Text Is Steerable, but Not by Simple Negation}

Positive natural-language requests consistently steer recommendations in the
intended direction. Across all eight evaluated genres and moods, requests such
as ``more jazz'' move the top-ranked recommendations toward the requested
genre or mood ($p<0.0001$ for every direction), increasing mean Top-20 precision
from 56\% to 74\%. We compared appending user feedback to the existing taste profile
against regenerating the taste profile with an LLM. Appending the request produced recommendations that were semantically closer to the requested genre or mood
(0.031 versus 0.022), suggesting that preserving the existing profile is more
effective than regenerating it.

This limitation is equally informative. Requests such as ``no electronic
music'' move recommendations toward, rather than away from, the mentioned
genre. This occurs because the embedding primarily captures the semantic concept
``electronic music'' rather than the logical meaning introduced by the negation.
Consequently, simply re-embedding a negated request increases its similarity to
the very concept the user intends to avoid. Handling negative preferences
therefore requires explicit reasoning or filtering mechanisms rather than
embedding manipulation alone.

More complex forms of steering are also effective. Situational requests such as
``music for falling asleep'' or ``deep focus'' produce larger shifts than
genre- or mood-based requests (mean shift 0.069), and multi-turn interactions
progressively refine recommendations toward increasingly specific user
preferences. These findings reinforce the role of taste profiles as an interactive interface between users and foundation-model recommender systems, although confirming their impact on user satisfaction will require evaluation in deployed user-facing systems.

\subsection{Structural and Temporal Diagnostics}

Taste profiles and behavioral embeddings exhibit only modest structural agreement, indicating that they organize user preferences differently in representation space. Representational Similarity Analysis yields a Spearman correlation
of $\rho=0.286$, with 17.5\% nearest-neighbor overlap at $k=5$ and an Adjusted
Rand Index of 0.08. These metrics characterize the structural relationship
between the two representation spaces rather than quantifying the information captured by either representation. The stronger evidence for complementarity
comes from the downstream improvements observed when both representations are
used together.

Taste profiles also remain highly stable over short time horizons. Full-profile embedding cosine similarity is 0.954 after three days and 0.955 after seven days, with no significant differences across seven output measures (all Wilcoxon $p>0.05$). Approximately 95.0\% and 94.4\% of the corresponding inputs remain unchanged over the two intervals, indicating that the observed stability reflects consistent profile generation under largely unchanged inputs. Evaluating responsiveness to genuine preference change requires a different experimental setting and remains future work.

\section{Discussion and Conclusion}
\label{sec:discussion}

Our results suggest that taste profiles should be treated as contextual
interfaces rather than standalone substitutes for behavioral representations.
They must be evaluated through their consumers: because intrinsic dimensions
are not interchangeable and profile versions produce different downstream
trade-offs, no single judge score is sufficient for system selection. 
Behavioral embeddings remain stronger for compact prediction, whereas text provides semantic organization, inspectability, and an interface for explicit context; accordingly, our most consistent gains come from combining the two. Although both representations originate from the same behavioral signals, text introduces a different inductive bias through language-model knowledge and prompt design. This complementarity may diminish as behavioral encoders
incorporate richer metadata or language-model reasoning. Successful deployment requires more than text similarity: positive requests can steer the
representation, but negation, conflicts between stated and observed
preferences, and changing intent require explicit policies and signal
provenance. Similarly, refresh strategies must balance stability under
unchanged inputs with responsiveness to meaningful events, while safety must
be enforced at the representation layer because profile inferences are visible
to users and reused across systems.

These findings should be interpreted in light of several limitations. Taste profile generation is an unsupervised summarization problem with no canonical ground-truth profile. Although our evaluation suite
measures factuality, coverage, specificity, relevance, safety, behavioral
signal, and downstream utility, the absence of gold labels makes it better
suited to comparing system variants than establishing absolute profile quality.
Taste profiles are also a new product and representation surface, so mature
interaction patterns and long-term measures of user value are not yet
available, while proprietary data and infrastructure limit exact reproduction.
Deployment to a large population demonstrates operational feasibility, but
does not imply that every profile is equally accurate or that every consumer
benefits; each application still requires controlled evaluation and monitoring.
Within these bounds, Textual User Taste demonstrates how natural-language user
context can be generated, evaluated, refreshed, safeguarded, and served at
scale. Rather than superseding behavioral embeddings, taste profiles provide
a complementary, interpretable, and steerable interface between evolving user
context and foundation-model recommender systems.

\bibliographystyle{ACM-Reference-Format}
\bibliography{references}

\end{document}